\pdfoutput=1

\documentclass[11pt,a4paper]{article}
\usepackage[hyperref]{emnlp2020}
\usepackage{times}
\usepackage{latexsym}

\usepackage{subcaption}

\usepackage{microtype}

\aclfinalcopy 
\def\aclpaperid{1351} 

\title{On the importance of pre-training data volume for compact language models}

\author{Vincent Micheli \\
  Illuin Technology / EPFL \\
  \texttt{vincent.micheli@illuin.tech}
  \And
  Martin d'Hoffschmidt \\
  Illuin Technology \\
  \texttt{martin.dhoffschmidt@illuin.tech}
  \AND
  François Fleuret \\
  UNIGE \\
  \texttt{francois.fleuret@unige.ch}
  }

\date{}

\begin{document}
\maketitle
\begin{abstract}
Recent advances in language modeling have led to computationally intensive and resource-demanding state-of-the-art models. In an effort towards sustainable practices, we study the impact of pre-training data volume on compact language models. Multiple BERT-based models are trained on gradually increasing amounts of French text. Through fine-tuning on the French Question Answering Dataset (FQuAD), we observe that well-performing models are obtained with as little as 100 MB of text. In addition, we show that past critically low amounts of pre-training data, an intermediate pre-training step on the task-specific corpus does not yield substantial improvements.
\end{abstract}

\section{Introduction}

Over the past year, pre-trained language models have become the norm in Natural Language Processing. These large-scale Transformer-based \citep{attention} networks considerably advanced the state-of-the-art in language understanding \citep{bert} via a two-step process: self-supervised learning on a vast text corpus followed by fine-tuning on a specific downstream task. 

Following these advances, the ongoing trend has been to build bigger models with an ever-increasing amount of data \citep{roberta, t5, gpt2, gpt3}. However, pre-training models with billions of parameters over hundreds of gigabytes of text requires tremendous computational resources that only a few companies and institutions can afford. Besides, many languages and specific corpora (e.g. legal, scientific) are currently under-resourced. Hence, our goal is to explore model architectures and data volumes lowering the entry barrier to new research and practical applications. 

We conduct experiments on French corpora in order to release the first French compact language models and to illustrate the training process in another language than English. Furthermore, we consider the question answering task since compact models may find their purpose in low-latency/fault-tolerant information retrieval systems.

\section{Problem statement}
We intend to study the impact of pre-training data volume when training compact bidirectional Transformers \citep{bert}. We assume a scarce resources setting, both in terms of data and computing power. Two key aspects are explored:

\begin{itemize}
    \item The amount of pre-training data required to train high-performing compact language models.
    \item The importance of corpus-specific MLM before fine-tuning.
\end{itemize}

We use the French part of the OSCAR corpora \citep{oscar} for pre-training and the FQuAD dataset \footnote{\url{https://illuin-tech.github.io/FQuAD-explorer/}} \citep{fquad} for machine reading comprehension fine-tuning. Moreover, the models under consideration are based on the CamemBERT \citep{camembert} language model.

\section{Related work}

A wealth of work has recently been released \citep{survey} on compressing Transformer-based models \citep{attention, bert} through the pre-training of compact models \citep{wellread}, distillation \citep{distillation, tinybert, mobilebert}, pruning \citep{trainthencompress, structuredpruning, movementpruning, dropoutpruning} and quantization \citep{qbert, quantnoise}. Nevertheless, absolute performance is not the end goal of this study. Rather, we investigate the training process of compact models in the absence of larger ones to distillate or prune. Furthermore, \citet{movementpruning} acknowledge the difficulty of speeding up sparse models due to the absence of specialized hardware. Therefore, from an inference speed standpoint, it is currently preferable to train compact models.

Language models have been successfully pre-trained on domain-specific corpora \citep{scibert, biobert} and outperform their general-purpose counterparts on targetted downstream tasks. Still, training these models involved large datasets and computational resources out of reach for most.

Multilingual models \citep{bert, xlm, xlm-r} have been released to alleviate the need for language-specific pre-training. While they offer competitive results, they usually lag behind monolingual models and require larger architectures.

\citet{camembert} observed that large models did not improve on evaluation tasks when increasing the amount of pre-training data from 4 GB to 138 GB. They left as future work to question the need for large scale pre-training corpora with other model architectures and tasks.

\section{Datasets}

\subsection{OSCAR}

OSCAR \footnote{\url{https://oscar-corpus.com/}} \citep{oscar} is a large-scale multilingual open source collection of corpora obtained by language classification and filtering of the Common Crawl corpus \footnote{ \url{https://commoncrawl.org/about/}}. The whole French part amounts to 138 GB of text and it has already been used to train French language models \citep{camembert}. In this work, we only extract a sample of 4 GB of shuffled lines.

\subsection{FQuAD}

FQuAD \citep{fquad} is a recently introduced open source French native reading comprehension dataset. It consists of 60,000 questions and answers gathered on a set of 1,769 high-quality Wikipedia articles. In many aspects, it is the French equivalent of SQuAD 1.1 \citep{squad}. Given a question and a paragraph, the task consists in extracting from the paragraph the span of text answering the question.

We chose FQuAD as the fine-tuning dataset because it allows one to draw a direct parallel with its English counterpart \citep{fquad} and is one of the largest annotated French datasets. However, question answering is a notoriously difficult task for compact models \citep{structuredpruning}. While distillation has shown to improve their results on the GLUE benchmark \citep{glue} substantially, machine reading comprehension remains difficult to speed-up without incurring a significant drop in accuracy. 

\section{CamemBERT\textsubscript{SMALL}} \label{camembert-small}

\begin{table}
\centering
\begin{tabular}{lrl}
\hline \textbf{Model} & \textbf{Size} & \textbf{Time} \\ \hline
CamemBERT\textsubscript{SMALL} & 72 MB & 157 ms \\
CamemBERT\textsubscript{BASE} & 440 MB & 705 ms \\
CamemBERT\textsubscript{LARGE} & 1340 MB & 2376 ms \\
\hline
\end{tabular}
\caption{\label{inferencetime-table} Model size and inference time on an Intel Xeon 2.30GHz Quad core CPU with batch size 1 and max sequence length 384 tokens (average over 1000 samples).}
\end{table}

CamemBERT \citep{camembert} is a multi-layer bidirectional Transformer \citep{attention} with two architectures: base (12 layers, 768 hidden dimensions, 12 attention heads, 110M parameters) and large (24 layers, 1024 hidden dimensions, 16 attention heads, 355M parameters). It is very similar to RoBERTa \citep{roberta}. The main differences are the use of whole-word masking and SentencePiece tokenization \citep{sentencepiece} instead of subword-masking and byte-level Byte-Pair encoding \citep{bpe, gpt2}. RoBERTa itself improves upon BERT by aggregating several modifications on top of the original architecture such as removing the next sentence prediction task, dynamic masking, and training with larger batches on more data. 

We introduce CamemBERT\textsubscript{SMALL} \footnote{The pre-trained models are made available in the Hugging Face collection: \url{https://huggingface.co/illuin/lepetit}.}, a CamemBERT-based language model with a small architecture (12 layers, 256 hidden dimensions, 4 attention heads, 17M parameters). The main difference with the original CamemBERT lies in the use of subword-masking. Indeed, the authors later found out that whole-word masking had at best a marginal impact on downstream task performance.

Apart from inference speed and size considerations, two main factors explain this architectural choice:

\begin{itemize}
    \item This is the same architecture as ELECTRA\textsubscript{SMALL++} \citep{electra}, a recently released compact language model. Even though ELECTRA and CamemBERT differ in many regards (ELECTRA being trained as a discriminator rather than a generator), prior experiments conducted by \citet{electra} give us an acceptable set of hyperparameters when pre-training and fine-tuning the model.
    \item \citet{wellread}'s empirical results suggest that depth should be prioritized over width when pre-training compact models. 
\end{itemize} 

Table \ref{inferencetime-table} shows that CamemBERT\textsubscript{SMALL} is much smaller and faster than its larger siblings. In a plausible setup for question answering systems, it provides, respectively, a 4.5-fold and 15-fold inference speed-up compared to CamemBERT\textsubscript{BASE} and CamemBERT\textsubscript{LARGE} while being 6.2 and 18.8 times smaller.

\section{Experiments} \label{experiments}

\begin{table}
\centering
\begin{tabular}{lll}
\hline \textbf{Hyperparameter} & \textbf{Pre-train} & \textbf{Fine-tune} \\ \hline
Train steps & 200k & 30k \\
Warmup steps & 10k & 3k \\
Batch size & 128 & 32 \\
Learning rate & 1e-4 & 1e-4 \\
Adam $\beta_1$ & 0.9 & 0.9 \\
Adam $\beta_2$ & 0.999 & 0.999 \\
Weight decay & 0.01 & 0.0 \\
Max gradient norm & 1.0 & 1.0 \\
Dropout & 0.1 & 0.1 \\
Mask percent & 15 & n/a \\
Max sequence length & 512 & 384 \\
\hline
\end{tabular}
\caption{\label{hyperparameters-table} Pre-training and fine-tuning hyperparameters. In the corpus-specific MLM step, we take the same hyperparameters as in pre-training except that we decrease the number of steps to 2.5k and drop the warmup.}
\end{table}

Six overlapping subsets are built from the 4 GB OSCAR sample. They are denoted as OSC\textsubscript{10}, OSC\textsubscript{100}, OSC\textsubscript{500}, OSC\textsubscript{1000}, OSC\textsubscript{2000} and OSC\textsubscript{4000} (the numbers indicating the number of MB). We extract an additional 10 MB sample from the corpus, which serves as a validation set for the self-supervised pre-training task. On the other hand, FQuAD consists of a train/dev split of 50,741 and 5,668 question/context pairs.

For each OSCAR subset, we pre-train a CamemBERT\textsubscript{SMALL} model with the standard masked language modeling (MLM) objective. 
Then we fine-tune the pre-trained models on the question answering task with the same span prediction method as BERT \citep{bert}. Between those two steps, an optional MLM step over the FQuAD train set is included.

Table \ref{hyperparameters-table} shows the pre-training, intermediate MLM (if any) and fine-tuning hyperparameters. 
Fine-tuning being a brittle process \citep{seedfinetuning}, fine-tuning results are averaged over 3 seeds. 

The experiments described were implemented using Hugging Face's Transformers library \citep{huggingface} and were conducted on an NVidia V100 16 GB.

\section{Analysis}

\citet{camembert} observed that complex downstream tasks may require more pre-training steps. Since for each OSCAR subset the validation loss is still slowly decreasing after 200k steps, we assume that training longer might increase performance on the difficult question answering task. On the other hand, corpus-specific MLM fine-tuning quickly converged for all models. Table \ref{results-table} reports the entirety of the results.

\begin{table*}
\begin{subtable}{.5\textwidth}
\centering
\begin{tabular}{lll}
\hline \textbf{Subset}  & \textbf{Perplexities} & \textbf{F1 score} \\ \hline
OSC\textsubscript{10}  & 45.20 / 43.34 & 58.18 (0.60) \\
OSC\textsubscript{100}  & 14.22 / 11.91 & 68.50  (0.25) \\
OSC\textsubscript{500}  & 12.75 / 10.58 & 69.50  (0.41)\\
OSC\textsubscript{1000}  & 12.56 / 10.57 & 69.35  (0.64)\\
OSC\textsubscript{2000}  & 12.45 / 10.41 & 70.96  (0.66)\\
OSC\textsubscript{4000}  & 12.49 / 10.35 & 69.76  (0.61)\\
\hline
\end{tabular}
\caption{\label{results-table-without} Without MLM fine-tuning.}
\end{subtable}
\hfill
\begin{subtable}{.5\textwidth}
\centering
\begin{tabular}{lll}
\hline \textbf{Subset}  & \textbf{Perplexities} & \textbf{F1 score} \\ \hline
OSC\textsubscript{10}  & 40.31 / 18.95 & 62.33  (0.58) \\
OSC\textsubscript{100}  & 16.35 / 9.41 & 69.04  (0.16) \\
OSC\textsubscript{500}  & 15.09 / 8.77 & 70.25  (0.43) \\
OSC\textsubscript{1000}  & 14.74 / 8.83  & 69.84  (0.27) \\
OSC\textsubscript{2000}  & 14.72 / 8.75 & 70.71  (0.08) \\
OSC\textsubscript{4000}  & 14.90 / 8.68 & 69.84  (0.79) \\
\hline
\end{tabular}
\caption{\label{results-table-with} With MLM fine-tuning.}
\end{subtable}
\caption{\label{results-table} Dev OSCAR / FQuAD perplexities and FQuAD F1 score (average token overlap between predicted and ground truth answers) for each pre-training subset.}
\end{table*}

\subsection{How much data does one need to pre-train a compact language model?}

As we increase the amount of pre-training data, perplexity on the OSCAR dev set decreases in every instance but one (OSC\textsubscript{4000}). Nevertheless, aside from OSC\textsubscript{10}, discrepancies are small and the models show almost identical learning curves. OSC\textsubscript{10} is underperforming in terms of MLM perplexity and question answering F1 score when compared to larger subsets. However, past this smallest dataset, pre-training data volume does not exhibit any strong monotonic relationship with downstream performance. The only OSCAR subset displaying a noticeable performance gap is OSC\textsubscript{2000}, with a +2.46 average F1 score increase over OSCAR\textsubscript{100}. For anchoring, a randomly initialized CamemBERT\textsubscript{SMALL} model "fine-tuned" directly on the FQuAD train set achieves an F1 score of only 17.76, i.e. 40 F1 points less than OSC\textsubscript{10}. This result indicates that even if a small amount of pre-training data is available, one should not neglect that step.
Regarding larger architectures, CamemBERT\textsubscript{BASE} and CamemBERT\textsubscript{LARGE} models from \citet{camembert} obtain an F1 score of 88 and 92, respectively, after fine-tuning.

Due to computational constraints, we could not investigate smaller or larger datasets as well as a prolonged pre-training phase. It could be the case that for a 200k pre-training steps budget, data volume is not the bottleneck. In fact, additional training steps may be even more beneficial for larger datasets. Nonetheless, a preliminary experiment pushing the pre-training phase of CamemBERT\textsubscript{SMALL} on OSC\textsubscript{2000} to 300k steps revealed that while the MLM loss decreased, the F1 score on the downstream task did not improve.

\subsection{Is corpus-specific MLM beneficial?}

Again, we observe a contrast between OSC\textsubscript{10} and larger subsets. OSC\textsubscript{10} is the only pre-training dataset significantly improving on the downstream task (+4.15 F1) and experiencing a decrease in perplexity on both pre-training and fine-tuning data when complemented with an intermediate MLM step. However, this corpus-specific MLM step is not truly intermediate since FQuAD contexts contain 10MB of raw text. This implies a 2-fold increase in pre-training data rather than a specific domain adaptation step. Therefore, we turn our focus to larger subsets for the rest of this analysis.

In these cases, MLM fine-tuning results in a net FQuAD perplexity decrease at the cost of an OSCAR perplexity increase. Domain shift may be the root cause of this trade-off. Indeed, as there exists a mismatch between pre-training and fine-tuning sets, the language model has to adapt to the specificity of descriptive paragraphs. In addition, perplexity is higher on the OSCAR dev set than on the FQuAD one. This is most likely due to the difficulty of predicting masked words in an heterogeneous web-crawled dataset compared to a set of high quality Wikipedia articles. 

For every pre-training subset but one (OSC\textsubscript{2000}), MLM fine-tuning induced a slight F1 score increase on the downstream task. However, these gains are marginal with at most a +0.75 average F1 score increase in the case of OSC\textsubscript{500}. Additional experiments are required to consolidate these findings, especially on larger task-specific datasets such as scientific or legal corpora. In those instances, a greater domain shift would probably legitimate an intermediate MLM fine-tuning step.

\section{Conclusion}

We investigated the importance of pre-training data volume when training compact Transformer-based models. We made the observation that 100 MB
of raw text are sufficient to reach similar performance as with larger datasets on a question answering task, and that corpus-specific self-supervised learning does not bring significant improvements on that particular problem. These preliminary results pave the way for further experiments with other language models, various architectures and new downstream tasks.

\section*{Acknowledgments}

We gratefully thank Quentin Heinrich for his reviewing and helpful discussions. We also thank Illuin Technology for its technical support and funding.

\newpage

\bibliographystyle{acl_natbib}
\bibliography{paper}

\end{document}